\documentclass[10pt,twocolumn,letterpaper]{article}

\usepackage{iccv}              
\usepackage{lipsum}
\usepackage{graphicx}
\def\1n{\mathbf{1}_n}
\def\0{\mathbf{0}}
\def\1{\mathbf{1}}

\definecolor{pink}{rgb}{0.9,0.5,0.5}
\definecolor{purple}{rgb}{0.5, 0.4, 0.8}   
\definecolor{gray}{rgb}{0.3, 0.3, 0.3}
\definecolor{mygreen}{rgb}{0.2, 0.6, 0.2}

\definecolor{greena}{rgb}{0.4, 0.5, 0.1}

\definecolor{bluea}{rgb}{0, 0.4, 0.6}

\definecolor{reda}{rgb}{0.6, 0.2, 0.1}

\newcommand{\cm}[1]{}

\newcommand{\myheading}[1]{\vspace{1ex}\noindent \textbf{#1}}

\newif\ifshowsolution
\showsolutiontrue

\ifshowsolution

\else

\fi

\definecolor{iccvblue}{rgb}{0.21,0.49,0.74}
\usepackage[pagebackref,breaklinks,colorlinks,allcolors=iccvblue]{hyperref}

\newcommand{\Approach}{CSD-VAR}

\def\paperID{XXXX} 
\def\confName{ICCV}
\def\confYear{2025}

\title{\Approach: Content-Style Decomposition in Visual Autoregressive Models}

\author{
Quang-Binh Nguyen\textsuperscript{1} \quad
Minh Luu\textsuperscript{2} \quad
Quang Nguyen\textsuperscript{1} \\
Anh Tran\textsuperscript{1} \quad
Khoi Nguyen\textsuperscript{1} \\
\textsuperscript{1}Qualcomm AI Research\textsuperscript{\dag}\quad 
\textsuperscript{2}MovianAI \\
{\tt\small \{binhnq, quanghon, anhtra, khoi\}@qti.qualcomm.com, v.minhlnh@vinai.io} \\
}

\definecolor{mydarkblue}{rgb}{0,0.08,1}
\definecolor{mydarkgreen}{rgb}{0.02,0.6,0.02}
\definecolor{myred}{rgb}{1.0,0.0,0.0}
\definecolor{myred2}{rgb}{0.7,0.1,0.1}
\definecolor{mydarkblue2}{rgb}{0.05,0.1,0.7}
\definecolor{mypurple}{rgb}{111,0,255}
\definecolor{mypurple2}{rgb}{111,0,111}

\newcommand\blfootnote[1]{%
  \begingroup
  \renewcommand\thefootnote{}\footnote{#1}%
  \addtocounter{footnote}{-1}%
  \endgroup
}

\begin{document}
\maketitle

\blfootnote{\textsuperscript{\dag} {\fontsize{7.5}{9}\selectfont \mbox{Qualcomm~AI~Research~is~an~initiative~of~Qualcomm~Technologies,~Inc.}}}

\begin{abstract}
Disentangling content and style from a single image, known as content-style decomposition (CSD), enables recontextualization of extracted content and stylization of extracted styles, offering greater creative flexibility in visual synthesis. While recent personalization methods have explored the decomposition of explicit content style, they remain tailored for diffusion models. Meanwhile, Visual Autoregressive Modeling (VAR) has emerged as a promising alternative with a next-scale prediction paradigm, achieving performance comparable to that of diffusion models. In this paper, we explore VAR as a generative framework for CSD, leveraging its scale-wise generation process for improved disentanglement.  To this end, we propose CSD-VAR, a novel method that introduces three key innovations: (1) a scale-aware alternating optimization strategy that aligns content and style representation with their respective scales to enhance separation, (2) an SVD-based rectification method to mitigate content leakage into style representations, and (3) an Augmented Key-Value (K-V) memory enhancing content identity preservation. To benchmark this task, we introduce CSD-100, a dataset specifically designed for content-style decomposition, featuring diverse subjects rendered in various artistic styles. Experiments demonstrate that CSD-VAR outperforms prior approaches, achieving superior content preservation and stylization fidelity.
\end{abstract}
\vspace{-5pt}    
\section{Introduction}
\label{sec:intro}

\begin{figure}[t]
    \centering
   \includegraphics[width=\linewidth]{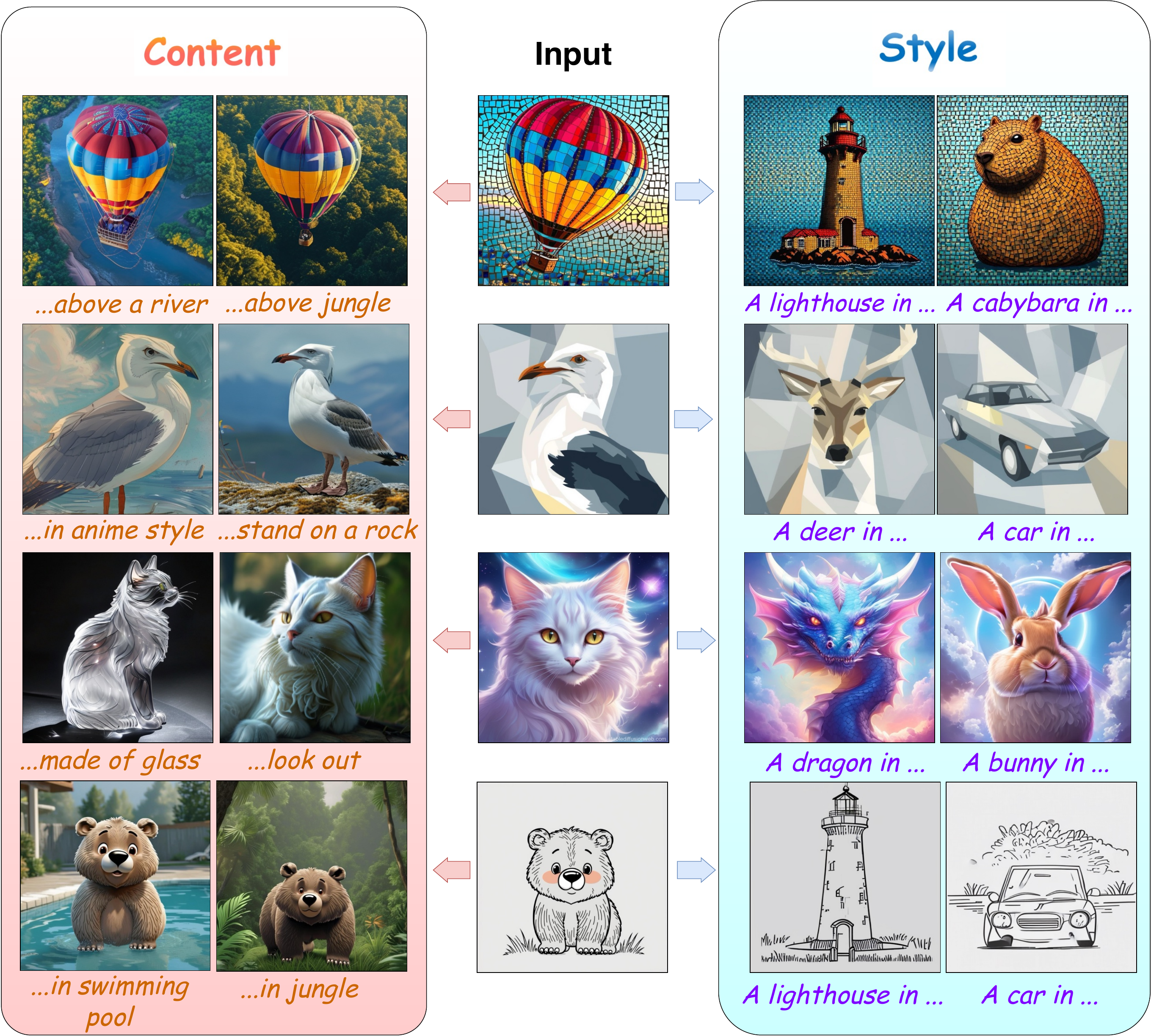}
    \caption{Given a single input image, our framework separates content and style, enabling flexible recontextualization and stylization to generate new images across diverse contexts.}
	\label{fig:teaser}
    \vspace{-10pt}
\end{figure}

The challenge of disentangling content and style, or \textbf{content-style decomposition (CSD)}, from a single image can be framed as a dual \textit{personalization problem}: one that extracts the subject’s structure and details (content) while separately capturing the artistic technique (style). Successfully addressing this problem enables two key applications: \textit{recontextualization}, where a subject is adapted to different visual environments, and \textit{stylization}, where an extracted style is applied to new subjects.
For instance, consider an artist working from a single reference -- a balloon illustration rendered in a unique artistic style, illustrated in \cref{fig:teaser} (first row). Separating its content and style would allow the artist to depict the same balloon in a realistic setting or apply the illustration’s stylistic elements to new subjects, such as landscapes or portraits. This separation fosters greater creative flexibility, enabling novel compositions and cross-domain visual transformations.

In text-to-image personalization, \textbf{textual inversion} \cite{textualinversion} optimizes text embeddings to capture specific concepts within the text encoder space while keeping model parameters frozen. Many subsequent methods~\cite{inspiration_tree,lego,cusconcept,conceptexpress,reversion,zhang2024compositional,gradientfree,mcpl} have extended textual inversion to better preserve object identity and enhance performance across various tasks beyond encoding only a subject’s appearance. However, these approaches do not explicitly separate content and style.  
More recent methods, such as B-LoRA \cite{blora} and UnZipLoRA \cite{unziplora}, address this limitation by decomposing an image into distinct content and style representations. Nevertheless, these techniques are specifically designed for diffusion-based models, and no prior work has explored their application to other generative models, such as Autoregressive (AR) models. 
Recently, AR models~\cite{llamagen,janus,chameleon,anole,mars,showo,var,hart,star,voronov2024switti,infinity} have emerged as a compelling alternative to diffusion models, offering comparable generative capabilities while maintaining high efficiency.  
A notable variant, \textbf{Visual Autoregressive Modeling (VAR)}~\cite{var,hart,star,voronov2024switti,infinity}, introduces a next-scale prediction paradigm, where generation starts with a $1\times1$ token map and progressively expands into a sequence of multi-scale token maps (e.g., $2\times2$, …), increasing in resolution. This motivates our investigation into how content and style representations can be effectively learned within VAR.  

However, directly applying textual inversion to VAR by optimizing separate style and content embeddings results in suboptimal representations due to the strong entanglement between these attributes. The inherent coupling of content and style makes simple text prompt guidance insufficient for effective decomposition. First, we empirically observe that early scales primarily encode style, while later scales capture content information. Building on this insight, we propose a \textbf{scale-aware alternating optimization} strategy, where content and style embeddings are optimized at their corresponding scales in an alternating manner to ensure better disentanglement. Second, to further strengthen content-style separation, we introduce a singular value decomposition \textbf{(SVD)-based rectification method}, which explicitly enforces independence between the content and style spaces, reducing mutual interference and enhancing the quality of learned representations. Third, relying solely on textual embeddings proves inadequate for capturing complex concepts and styles, particularly in intricate cases. To address this limitation, we introduce \textbf{augmented Key-Value (K-V) memory}, which serves as auxiliary storage for content and style attributes that textual inversion alone fails to capture. These augmented K-V matrices not only improve content-style disentanglement but also enhance identity preservation, ensuring more faithful and expressive representations.

To our knowledge, no publicly available dataset with quantifiable benchmarks exists for content-style decomposition (CSD). While existing datasets focus on either style transfer or content preservation, they do not fully meet the requirements for evaluating CSD, prompting us to introduce CSD-100, a dataset of 100 images designed specifically for this task. On \textbf{CSD-100}, our \Approach~outperforms state-of-the-art by a large margin in all common metrics. 

In summary, our main contribution includes:

 \begin{itemize}
    \item We are the first to explore VAR-based personalization for content-style decomposition.
    \item We analyze scale-dependent representations in VAR and propose a scale-aligned optimization strategy to improve content-style disentanglement.
    \item We introduce an SVD-based constraint to enforce orthogonality between content and style representations.
    \item We propose augmented K-V memory to enhance disentanglement and better preserve subject identity.
    \item We propose a new dataset, CSD-100, specifically designed for the CSD with various styles and subjects.
\end{itemize}

\section{Related Work}

\myheading{Text-to-image generative models.} 
Early text-to-image generation methods built on \textbf{GANs} \cite{gan}, conditioning on textual inputs to map embeddings into image space, but suffered from training instability and mode collapse. The introduction of \textbf{diffusion models} (DMs) \cite{ramesh2021zero,rombach2022high,saharia2022photorealistic} improved image quality and diversity by leveraging large-scale image-text datasets and iterative denoising.  
However, DMs are computationally expensive due to slow sequential inference. To mitigate this, recent works \cite{luo2023latent,meng2023distillation,salimans2022progressive} distill multi-step teacher models into efficient few-step student networks, with some methods \cite{swiftbrushv2,liu2023instaflow,nguyen2024swiftbrush,yin2025improved} achieving one-step text-to-image generation.

\myheading{Autoregressive (AR) models} offer an alternative to diffusion-based approaches. Earlier AR models generated images sequentially via next-token prediction \cite{llamagen,janus,chameleon,anole}, but their step-by-step nature made high-resolution generation slow. Recently, next-scale prediction AR has emerged as a more efficient paradigm, generating images progressively from low to high resolution \cite{var,hart,star,voronov2024switti,infinity}, refining details at each scale while attending to previous ones.  
This approach has been successfully applied to text-to-image generation: STAR \cite{star} and Switti \cite{voronov2024switti} use cross-attention for text conditioning, HART \cite{hart} improves VAR \cite{var} with a diffusion model for continuous error residuals, and Infinity \cite{infinity} enables fine-grained generation with bitwise token prediction. While our method is compatible with any next-scale AR model, we use Switti \cite{voronov2024switti} and Infinity \cite{infinity} as our backbones due to their available training and inference code.

\myheading{Image personalization.}  
Personalization methods adapt a pretrained model to integrate a specific concept from a given set of images, typically by optimizing specialized \textbf{text embeddings}~\cite{textualinversion} or fine-tuning \textbf{model weights}~\cite{dreambooth}. While fine-tuning improves reconstruction fidelity, it requires substantial memory and often leads to overfitting. To mitigate these issues, Parameter-Efficient Fine-Tuning (PEFT) techniques~\cite{peft,lora,multiconcept,mixofshow} enhance efficiency while reducing memory overhead. 
Textual inversion methods have shown strong generalization across various tasks~\cite{reversion,zhang2024compositional,inspiration_tree,mcpl,pez,tokenverse}, extending beyond appearance capture due to the flexibility of text embedding manipulation. PEZ~\cite{pez} optimizes hard text prompts to capture diverse targets, while Reversion~\cite{reversion} introduces contrastive learning to steer embeddings into relational space. TokenVerse~\cite{tokenverse} further explores modulation space in DiT-based models, demonstrating that fine-tuned offset text embeddings can encode various concepts, including objects, accessories, materials, and lighting.  
However, these methods are designed for diffusion-based models, and directly applying them to VAR models does not yield effective results. In this work, we focus on textual inversion for content-style decomposition within the VAR framework, introducing techniques that explicitly leverage VAR’s multi-scale generation process for improved disentanglement.


\begin{figure*}[t!]
	\centering
	\includegraphics[width=\linewidth]{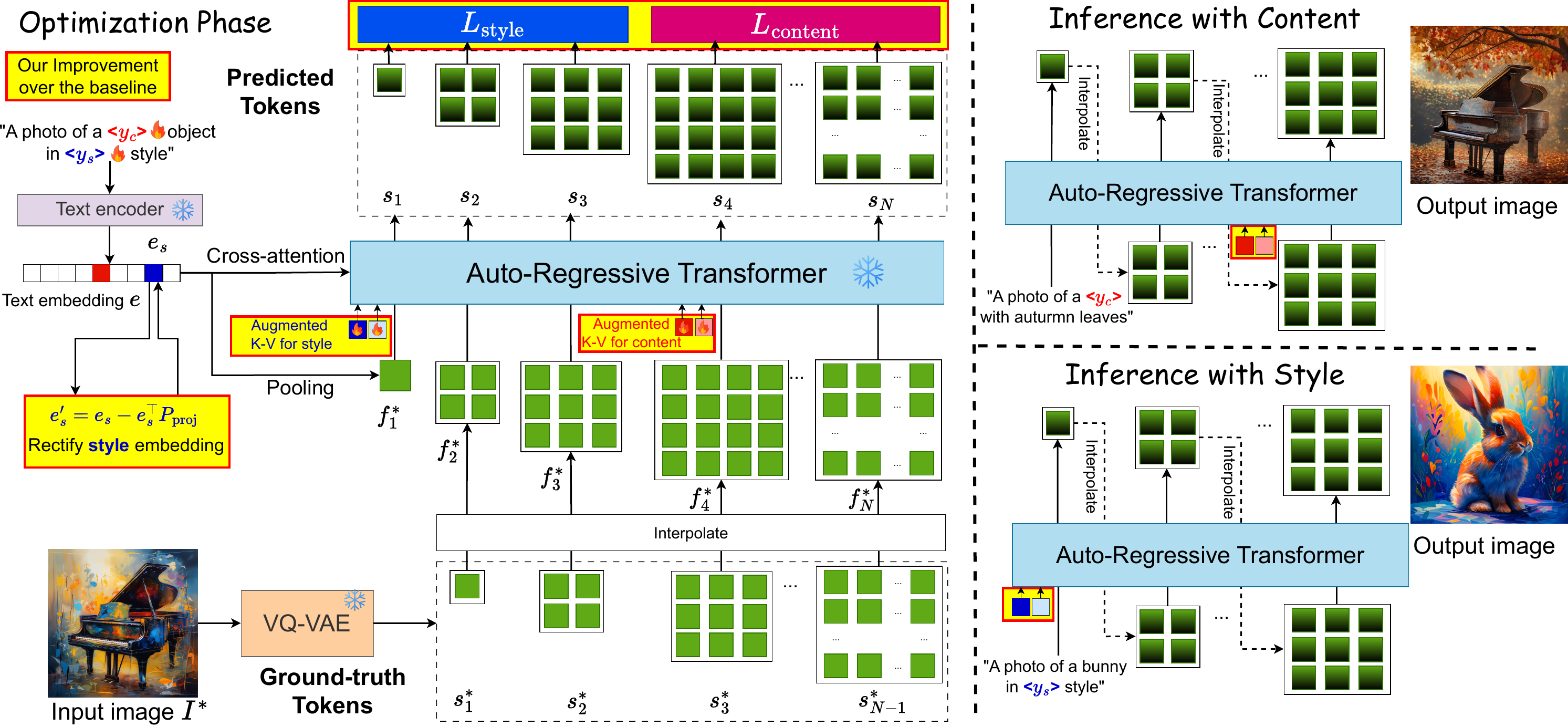}
	\caption{\textbf{Overview of our \Approach.} During optimization (left), a content-style prompt $\mathbf{y}$, \texttt{"A photo of a <$y_c$> object in <$y_s$> style"}, is encoded into text embeddings $\mathbf{e}$. The rectified style embedding $e_s$ reduces content leakage, while ground-truth scale-wise tokens from VQ-VAE are interpolated for next-scale prediction. Augmented K-V memories are prepended at specific scales before feeding into the autoregressive transformer. The model is trained with scale-wise cross-entropy losses, alternating optimization of content and style embeddings.  
At inference (right), style or content K-V memories are prepended based on the prompt before predicting tokens.
}
	\label{fig:main_diagram}
    \vspace{-10pt}
\end{figure*}

\myheading{Content-Style decomposition.}  
The task of concept extraction has been actively researched for years~\cite{breakascene,inspiration_tree,mcpl,attentioncali,omniprism}, with recent advances extending beyond simple subject representation. More recently, UnzipLoRA~\cite{unziplora} introduced content-style decomposition, emphasizing the simultaneous optimization of content and style representations. This ensures that extracted content remains adaptable across diverse contexts while preserving fidelity, and the style representation generalizes across subjects without content leakage.  
As a \textbf{fine-tuning approach}, UnzipLoRA achieves this by jointly learning two LoRA modules for content and style, employing prompt separation and block-wise sensitivity analysis. B-LoRA~\cite{blora} fine-tunes only two sets of sensitive layers for implicit content-style decomposition. U-VAP~\cite{uvap} enables fine-grained personalization by learning local subject attributes but is unsuitable for global characteristics like style and relies on LLM-guided data augmentation, limiting scalability.  
In \textbf{textual inversion}, Inspiration Tree~\cite{inspiration_tree} learns hierarchical subconcepts, though disentanglement remains unpredictable. Lego~\cite{lego} extends textual inversion beyond single-concept appearance, while ConceptExpress~\cite{conceptexpress} enables unsupervised concept extraction but is restricted to spatially localized attribute masks, making it ineffective for capturing global style.  
 
While previous methods focus primarily on fine-tuning diffusion models, our approach explores content-style decomposition within an autoregressive framework, leveraging the multi-scale generation process of VAR models.

\section{Preliminaries}
\label{sec:pre}

\myheading{Visual Autoregressive Modeling (VAR).} Unlike conventional next-token prediction autoregressive methods, VAR~\cite{var} introduces a novel paradigm in visual autoregressive modeling, shifting the focus from next-token prediction to next-scale prediction. Instead of generating tokens sequentially, VAR predicts an entire token map at each scale based on previously predicted scales. The process starts with a $1 \times 1$ token map, denoted as $s_1$, and progressively generates a sequence of multi-scale token maps $(s_1, s_2, \dots, s_K)$, increasing in resolution up to $K \times K$. 

VAR $\theta$, visualized in \cref{fig:main_diagram} (excluding the \colorbox{yellow!100}{yellow box}), models the following joint distribution to generate the token maps of different scales sequentially:
\begin{equation}
\label{equ:var_formula}
p({s}_{1}, {s}_{2},\dots,{s}_{K}) = \displaystyle\prod_{k=1}^{K}p_{\theta}( {s}_{k}| {s}_{1}, {s}_{2},\dots,{s}_{k-1}; c),
\end{equation}
where $c$ is the condition such as object class, ${s}_{k} \in {[V]}^{{h}_{k} \times {w}_{k}}$ is the token map at scale $k$, with size ${h}_{k} \times {w}_{k}$, conditioned on the previously generated scales $({s}_{1}, {s}_{2},\dots,{s}_{k-1})$. Each token in ${s}_{k}$ is an index from the VQ-VAE codebook with $V$ vocabularies, trained through multi-scale quantization and shared across scales.
In practice, to maintain a similar number of input and output tokens at the current scale, as required by the autoregressive transformer, the output token at previous scale is interpolated to yield $f^*_{k}$, before being fed back into VAR as input for the next scale  $k$, or:
\begin{gather}
\label{equ:var_formula_with_f}
p({s}_{1}, {s}_{2},\dots,{s}_{K}) = \displaystyle\prod_{k=1}^{K}p_\theta( s_{k}| {f}_{1}, {f}_{2},\dots,{f}_{k}; c),\\
\text{ where } f_k = \text{Interpolate}(s_{k-1}, k).
\end{gather}

\begin{figure}[t]
	\centering
	\includegraphics[width=\linewidth]{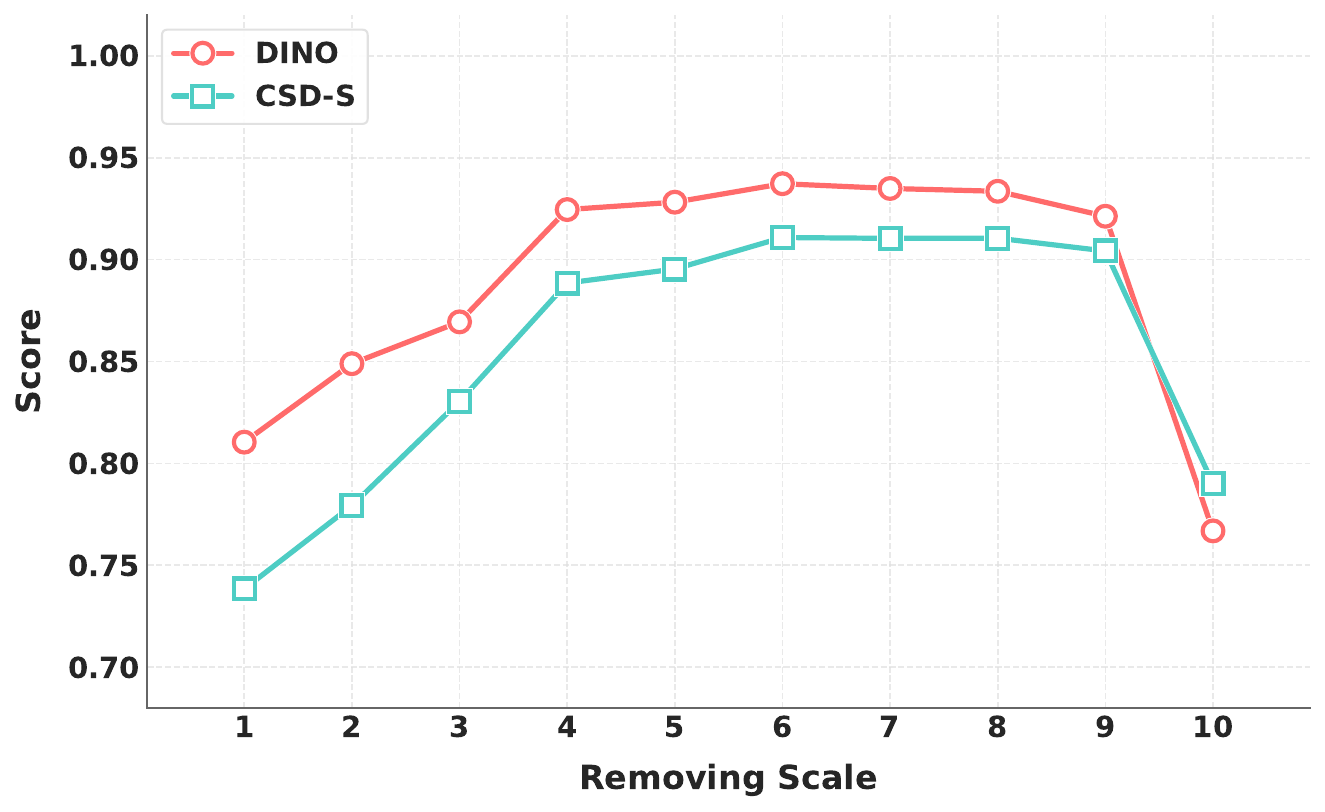}
	\caption{\bf Analysis of style-related scores across different scales.}
    \vspace{-10pt}
    \label{fig:4_style_score_analysis}
    \vspace{-5pt}
\end{figure}

To generate the final output image (\cref{fig:main_diagram}, Right), we interpolate all predicted token maps across scales $s_k$ to the final scale $K$, sum them together, and then pass the result through a decoder $\mathcal{D}$, as follows:
\begin{equation}
\label{equ:var_decode}
s_K = \displaystyle\sum_{k=1}^{K} \text{Interpolate}(s_k, K); \quad I = \mathcal{D}(s_K).
\end{equation}

\myheading{Teacher forcing training} (\cref{fig:main_diagram}, Left) is a widely used technique to accelerate the training of AR models by enabling parallel training across different scales. Instead of sequentially generating each scale and using its output as a condition for the next, this approach feeds ground-truth scales directly as conditions. This allows the model to learn more efficiently without waiting for the completion of previous scale predictions. In particular,
\begin{equation}
\label{equ:var_teacher_forcing}
p({s}_{1}, {s}_{2},\dots,{s}_{K}) = \displaystyle\prod_{k=1}^{K}p_\theta( s_{k}|f^*_1, {f}^*_{2},\ldots,{f}^*_{k}; c),
\end{equation}
where $\{f_k^*\}_{k=1}^K$ are the interpolated ground-truth scales, obtained by encoding the training image $I^*$ with the image encoder $\mathcal{E}$, e.g., VQ-VAE \cite{VQVAE}, or:
\begin{equation}
\label{equ:var_encode}
\{s_k^*\}_{k=1}^{K}= \mathcal{E}(I^*); \quad f^*_k = \text{Interpolate}(s^*_{k-1}, k).
\end{equation}

The training loss is the cross-entropy loss between predicted and ground-truth tokens at all scales.
\begin{equation}
\label{equ:var_loss}
\mathcal{L} = \displaystyle\sum_{k=1}^K \mathcal{L}_k = \displaystyle\sum_{k=1}^{K} - s^*_k \log p_\theta( s_{k}| {f}_{1}^*, {f}^*_{2},\dots,{f}^*_{k}; c).
\end{equation}

\myheading{Text-to-image VAR} extends the VAR framework for text-to-image synthesis. Specifically, we adopt Switti~\cite{voronov2024switti} and Infinity~\cite{infinity} as representative backbones for developing our approach due to their publicly available training and inference code. Given a text prompt $\mathbf{y}=\{y_i\}_{i=1}^{N}$ where $N$ is the number of words in the text prompt, we first encode it using both OpenCLIP ViT-bigG \cite{openclip} and CLIP ViT-L~\cite{clip}. The resulting embeddings $\mathbf{e}=\{e_i\}_{i=1}^{N}$ are then pooled to form the start token $f_1=\frac{1}{N}\sum_{i=1}^N e_i$ to predict the first scale $s_1$.
In each transformer block—comprising self-attention, cross-attention, and a feed-forward network (FFN)—the text embedding c is incorporated into the cross-attention layers via the standard attention mechanism~\cite{vaswani2017attention}.

\section{Our Approach}
\label{sec:method}
\myheading{Problem statement:} Given an input image $I^*$ containing a subject $y_c$ in style $y_s$, our objective is to disentangle its style and content into two distinct representations, enabling the generation of separate images: $I_c$, which accurately preserves the content of $I^*$, and $I_s$, which effectively captures its style. To this end, we explore the use of Visual Autoregressive Models (VAR) and leverage the pretrained text-to-image model, i.e., Switti \cite{voronov2024switti} and Infinity \cite{infinity}, for this task. 

\myheading{Our baseline:} A feasible approach is \textbf{Textual Inversion} \cite{textualinversion}, which optimizes the textual embeddings for content, $y_c$, and style, $y_s$, before the text encoder, without modifying Switti and Infinity’s pretrained weights -- thus preserving its generative capabilities. In particular, the prompt used for textual inversion is: \texttt{"A photo of a <$y_c$> object in <$y_s$> style"}, as shown in \cref{fig:main_diagram}. However, this simple baseline produces subpar results, as demonstrated in~\cref{tab:5_ablate_components} and~\cref{fig:5_qualitative_ablation}
. In the following section, we introduce \textbf{three key improvements} to enhance the effectiveness of Textual Inversion for text-to-image VAR.

\subsection{Scale-aware Alternating Optimization Strategy}

Examining the detail captured at each scale in VAR reveals correlations with content and style information. Inspired by this, we analyze how different scales contribute to style attributes (e.g., color, texture) and content structure (e.g., object shape, category, and fine-grained details).  

Using a set of 35 images with variations in content and style, we extract token maps across all scales, remove a specific scale’s tokens, and reconstruct the image. We then compute DINO~\cite{dino} and CSD-S~\cite{csd} scores to measure style similarity between the original image $I^*$ and the reconstructed image $I$, as these metrics are widely adopted for accessing style similarity~\cite{stylealign,rout2024rbmodulation}. As shown in \cref{fig:4_style_score_analysis}, removing smaller-scale token maps ($k = \{1,2,3\}$) and the final scale ($k = 10$) significantly affects style.  
Based on these findings and the assumed disentanglement of content and style, we categorize scales into two groups: the style-related group ($S_{\text{style}} = \{1,2,3,10\}$) and the content-related group ($S_{\text{content}} = \{4, \dots, 9\}$). The losses for optimizing the style embedding $y_s$ and content embedding $y_c$ are:
\begin{gather}
\label{equ:content_style_loss}
 \mathcal{L}_{\text{style}} = \displaystyle\sum_{k \in S_{\text{style}}}\mathcal{L}_k + \alpha \displaystyle\sum_{k' \in S_{\text{content}}}\mathcal{L}_{k'} \\
 \mathcal{L}_{\text{content}} = \displaystyle\sum_{k \in S_{\text{content}}}\mathcal{L}_k,
\end{gather}
where each $\mathcal{L}_k$ represents the cross-entropy loss at scale $k$, as defined in \cref{equ:var_loss}, and $\alpha \in (0,1)$ controls the influence of larger-scale token maps on style, ensuring that certain style attributes not fully captured by lower scales are retained.  
To further improve content-style disentanglement, we introduce an alternating optimization strategy, where content and style embeddings are optimized in separate iterations. This prevents gradient mixing, ensuring a clearer separation between the two representations.

\begin{figure}[t!]
	\centering
	\includegraphics[width=1.0\linewidth]{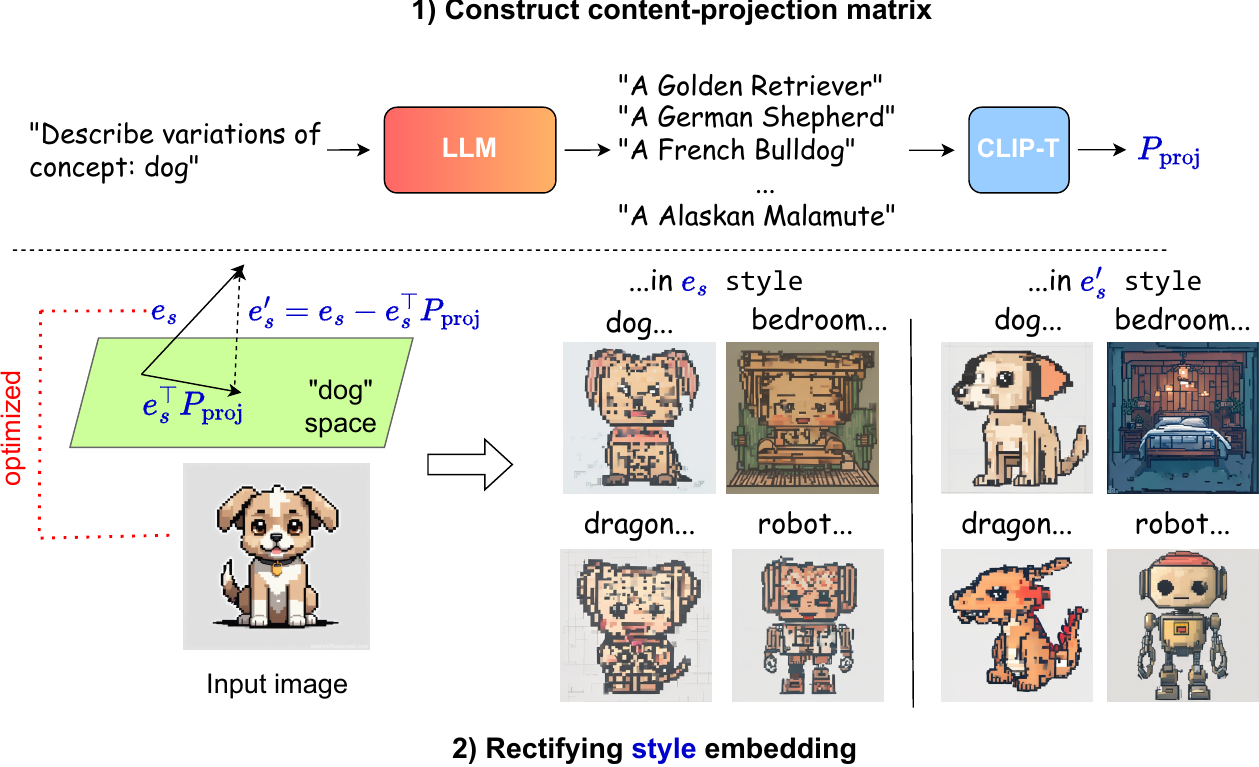}
	\caption{Style embedding rectification and examples}
	\label{fig:4_rectify}
    \vspace{-15pt}
\end{figure}

\subsection{SVD-based Style Embedding Rectification}
\label{sec:rectification}

While the style textual embedding $y_s$ is designed to capture texture details, its ability to attend to content scales (with a small $\alpha$ in \cref{equ:content_style_loss}) allows residual content information in smaller-scale token maps, leading to content leakage into the style embedding. This contamination results in an imperfect style representation, as shown in \cref{fig:4_rectify}. To mitigate this, we propose an SVD-based rectification method that refines the original style embedding $e_s = \text{CLIP}(y_s)$ by removing content-related information during training.  

First, we construct a content-related subspace by using a large language model (LLM), such as Llama~\cite{llama} or ChatGPT~\cite{gpt4}, to generate variations or subconcepts of the target concept. For instance, the concept ``dog'' may have subconcepts like ``Golden Retriever,'' ``German Shepherd,'' and ``Bulldog,'' which we embed using a CLIP text encoder.  

Next, we extract the most relevant directions within the content subspace by applying singular value decomposition (SVD) to the matrix $M \in \mathbb{R}^{Q \times d}$, where the columns are the text embeddings of the $Q$ sub-concepts. The SVD decomposition is: $M = U \Sigma V^T$, 
where $\Sigma$ is a diagonal matrix of singular values, and $U, V$ contain singular vectors representing dominant directions in the content embedding space. We then select the top $r$ singular vectors (corresponding to the largest singular values) to construct the projection matrix:  
\begin{equation}
P_{\text{proj}} = V_r^\top V_r,
\end{equation}
where $V_r \in \mathbb{R}^{r \times d}$ is the truncated version of $V$ up to rank $r$ and $P_{\text{proj}}$ remains fixed during training for a given concept.  

Finally, we remove content-related information from the style embedding $e_s$ by projecting it onto $P_{\text{proj}}$ and subtracting the projected content as follows:  
\begin{equation}
e'_s = e_s - e_s^\top P_{\text{proj}}.
\end{equation}  
This ensures that $e'_s$ remains orthogonal to content-related variations, effectively preventing unintended subject leakage in the generated images. As shown in \cref{fig:4_rectify}, rectifying $e'_s$ significantly reduces content leakage.

\subsection{Augmented Key-Value (K-V) Memories}
With the two improvements above, content and style embeddings can be effectively separated. However, for complex content or style concepts, textual embeddings alone may be insufficient, leading to under-captured representations. To address, we introduce augmented K-V memory to enhance the information captured by textual embeddings.  

Specifically, we augment several blocks in the autoregressive transformer with $O$ pairs of K-V memories, inserted before the self-attention layers. These K-V memories, denoted as $\tilde{K}$ and $\tilde{V}$, are added just before the $K, V$ matrices at the first scale for style ($k=1$) and the fourth scale for content ($k=4$), respectively. Formally:
\begin{equation}
\label{equ:prefix_attn}
\operatorname{Attn}(Q, K, V; \tilde{K}, \tilde{V}) = \operatorname{Attn} \left( Q, \begin{bmatrix} \tilde{K} \\ K \end{bmatrix}, \begin{bmatrix} \tilde{V} \\ V \end{bmatrix} \right).
\end{equation}

\begin{figure}[t]
  \centering
  \begin{subfigure}{\linewidth}
    \centering
    \includegraphics[width=\linewidth]{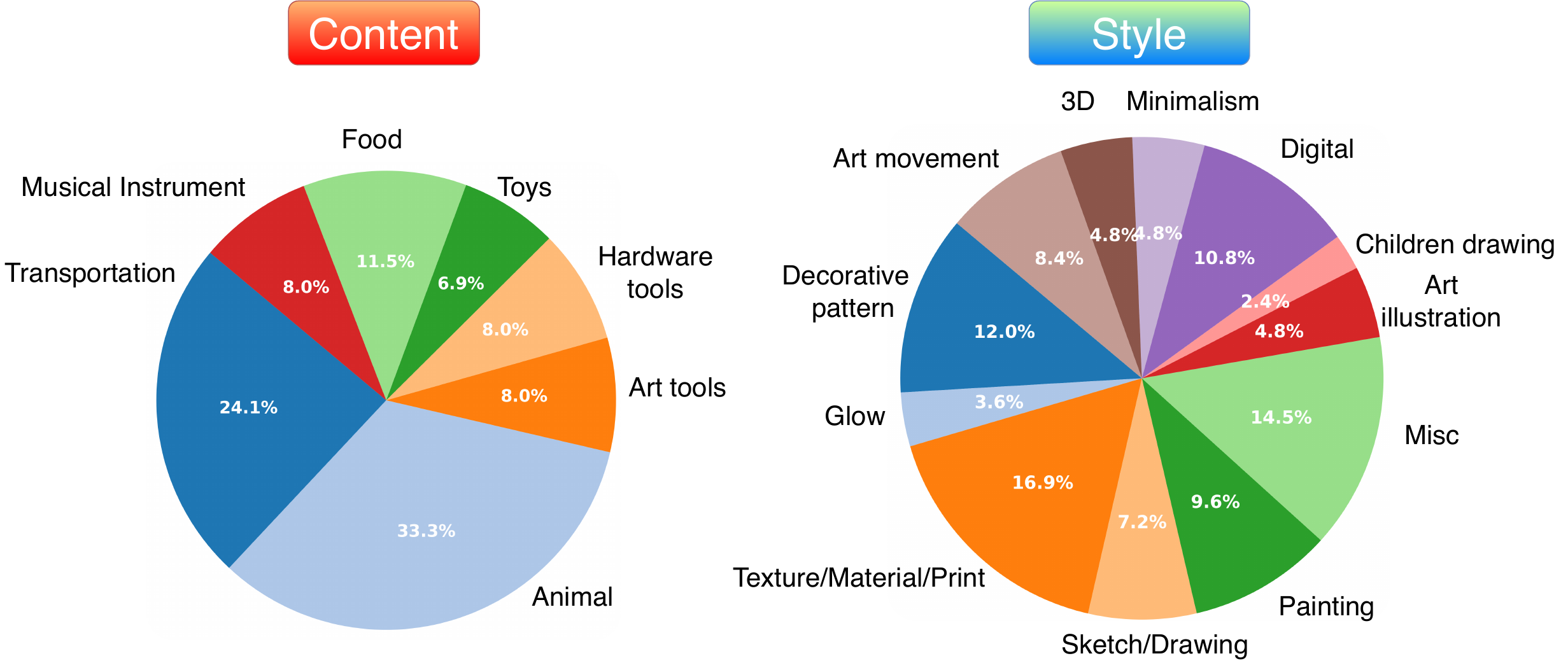}
    \caption{Distribution of content and style categories in CSD-100.}
  \end{subfigure}
  \begin{subfigure}{0.95\linewidth}
    \centering
    \includegraphics[width=\linewidth]{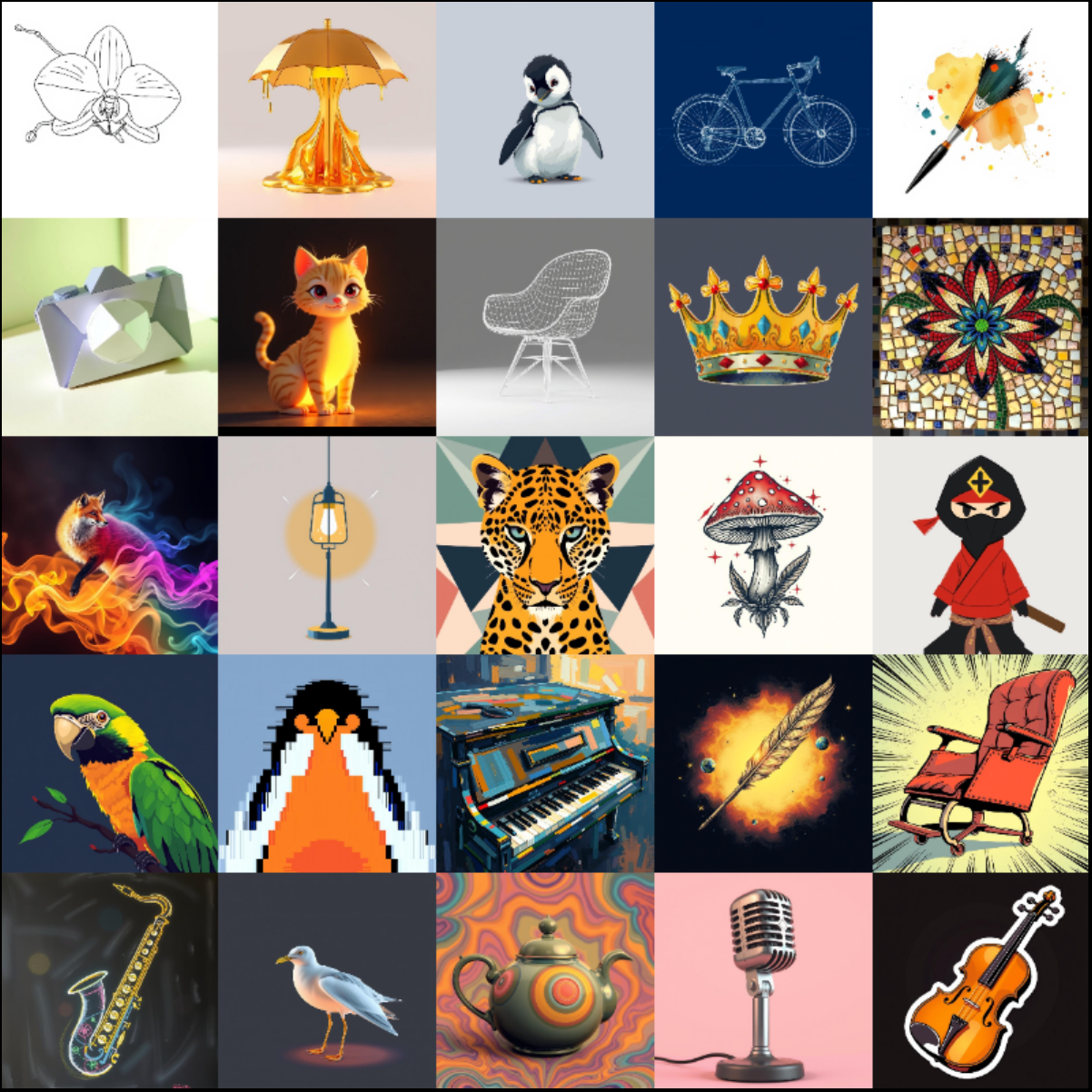}
    \caption{Samples from our CSD-100 dataset.}
  \end{subfigure}
  \vspace{-10pt}
  \caption{Statistics and samples of the CSD-100 dataset.}
  \label{fig:csd_100_overview}
\vspace{-15pt}
\end{figure}

\begin{figure*}[t!]
	\centering
	\includegraphics[width=\linewidth]{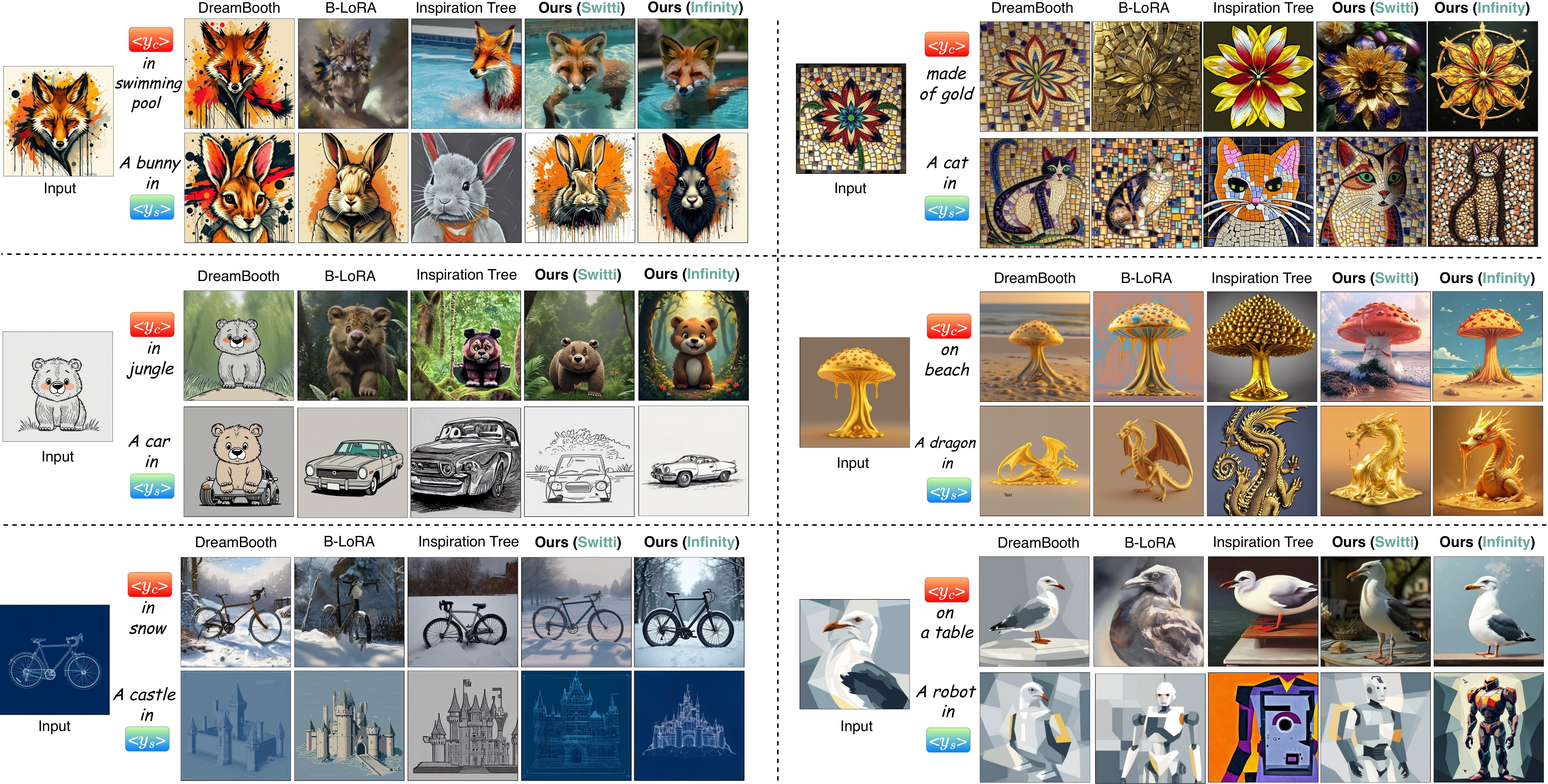}
        \vspace{-20pt}
	\caption{\textbf{Qualitative comparison with prior approaches on our CSD-100 dataset.}}
	\label{fig:5_qualitative_comparisons}
    \vspace{-10pt}
\end{figure*}

\vspace{-10pt}
\section{Our CSD-100 Dataset}
\label{sec:csd_100}

To our knowledge, no publicly available dataset provides quantifiable benchmarks for content-style decomposition (CSD). While existing datasets~\cite{styledrop,dreambooth,dreambenchplus,custom101} contain elements related to style transfer or content preservation, they do not fully meet the requirements for CSD evaluation.  
StyleDrop~\cite{styledrop} offers diverse style images but is relatively small and lacks sufficient instances where style is applied to salient objects. Similarly, personalization datasets such as DreamBooth~\cite{dreambooth}, and CustomConcept101~\cite{custom101} focus on content realism rather than explicit content-style representation. Whereas StyleAligned~\cite{stylealign} and RB-Modulation~\cite{rout2024rbmodulation} provide only benchmark prompts, limiting their usability for standardized evaluation.

To address this gap, we introduce \textbf{CSD-100}, a dataset of approximately 100 images designed to capture diverse content and styles for CSD evaluation. As shown in \cref{fig:csd_100_overview}, our CSD-100 spans a wide distribution of content and style, supporting robust evaluation. This section outlines the pipeline for generating and curating CSD-100.

\myheading{Data Collection and Curation Process.}  
We begin with the RB-Modulation~\cite{rout2024rbmodulation} prompt collection, containing approximately 400 content and 100 style concepts. To refine content categories, we remove overlapping terms and those related to landscapes, houses, humans, swings, and wheels, as they complicate content-style distinction. This filtering results in 180 content concepts while retaining all 100 style concepts.  
Next, we generate content-style image pairs using Flux 1.0~\cite{flux2024}, a text-to-image model, with the prompt \texttt{"A photo of <content> in <style>."}, producing around 18,000 images. To refine this, we develop a custom application for manual review, selecting the 10 most representative images per style while ensuring consistency by avoiding multiple objects in a single image. This reduces the dataset to 1,000 images.  

For further filtering, we use ChatGPT~\cite{gpt4} with a structured prompt to distill the dataset down to 100 high-quality images that best capture diverse content-style combinations. Illustrative examples are shown in \cref{fig:csd_100_overview}. These manual steps require \textbf{significant human effort}, and in the future, we plan to scale the dataset with more images, content concepts, and style categories.  

\myheading{Evaluation Protocol.}  
We curate 50 inference prompts (25 for content, 25 for style) and generate 10 images per prompt for each CSD-100 concept. This results in an extensive evaluation set of $100 \times 50 \times 10 = 50,000$ images, ensuring robust assessment. Details of inference prompts and dataset analysis are provided in the supplementary material.

\section{Experiments}

\subsection{Experimental Setup}

\myheading{Datasets:} To evaluate our method, we use the proposed CSD-100 dataset described in \cref{sec:csd_100}. Additionally, we construct a smaller validation set of 35 images for our ablation study. These images are curated from StyleDrop~\cite{styledrop}, B-LoRA~\cite{blora}, UnZipLoRA~\cite{unziplora}, and DreamBooth dataset \cite{dreambooth},
following the same evaluation protocol as CSD-100.

\myheading{Metrics:} We use the following metrics: CSD-C \cite{csd} and CLIP-I \cite{clip} for content alignment, CSD-S \cite{csd} and DINO \cite{dino} for style alignment, and CLIP-T \cite{clip} for text alignment.  For all metrics, higher values indicate better performance. 

\myheading{Implementation details: }
We build our approach on \textbf{Switti}~\cite{voronov2024switti} and \textbf{Infinty}~\cite{infinity} and optimize it using the Adam optimizer with a learning rate of $10^{-3}$ for 200 steps and a batch size of 1. The style loss coefficient is empirically set to  $\alpha = 0.1$  across all experiments. K-V memories are randomly initialized using the Xavier uniform initialization scheme \cite{xavier}. Training is conducted on a single A100 GPU. For generation variations in \cref{sec:rectification}, we generate 200 subconcepts per concept.

\begin{table}[t]
\small
    \centering
    \setlength{\tabcolsep}{1.5pt} 
    \begin{tabular}{lccccc}
        \toprule
        \bf Methods & \multicolumn{2}{c}{\bf Content Align} & \multicolumn{2}{c}{\bf Style Align} &\bf Text Align \\
        \cmidrule(lr){2-3} \cmidrule(lr){4-5} \cmidrule(lr){6-6}
        & \bf CSD-C& \bf CLIP-I & \bf CSD-S & \bf DINO & \bf CLIP-T \\
        \midrule
        DreamBooth-C~\cite{dreambooth} &0.594 &0.721 &-- &-- &0.271 \\
        DreamBooth-S~\cite{dreambooth} &-- &-- &0.537 &0.519 &0.289 \\
        B-LoRA~\cite{blora} &0.523 &0.592 &0.476 &0.346 &0.278 \\
        Inspiration Tree~\cite{inspiration_tree} &0.497 &0.575 &0.404 &0.353 & 0.257\\
        \midrule
        CSD-VAR (\textbf{\textcolor{teal}{Switti}}) &0.603 &0.754 &\textbf{0.564} &0.521 &\textbf{0.332} \\
        CSD-VAR (\textbf{\textcolor{teal}{Infinity}})&\textbf{0.660} &\textbf{0.795} &0.552 &\textbf{0.536} &0.319 \\
        \bottomrule
    \end{tabular}
    \vspace{-10pt}
    \caption{\bf Quantitative comparison on our CSD-100 dataset.}
    \label{tab:5_compare_sota}
    \vspace{-10pt}
\end{table}

\subsection{Comparison with Prior Approaches}

\begin{figure}[t!]
	\centering
	\includegraphics[trim={0 0 0 0},clip,width=\linewidth]{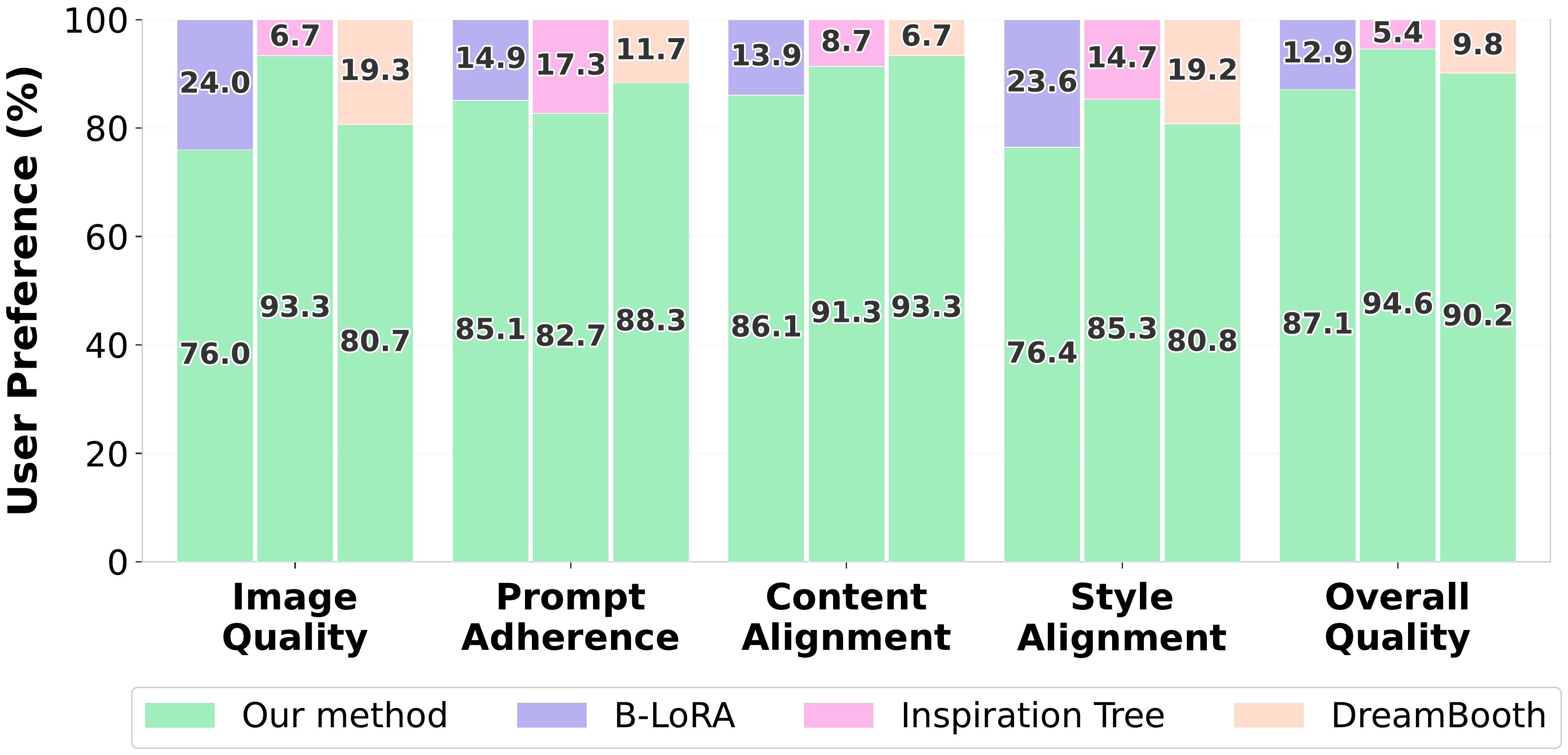}
        \vspace{-20pt}
	\caption{\textbf{User preference study on key criteria for content-style decomposition with 100 participants.}}
	\label{fig:userstudy}
    \vspace{-15pt}
\end{figure}

\myheading{Qualitative comparison.} To assess the effectiveness of CSD, we compare our method against DreamBooth~\cite{dreambooth}, B-LoRA~\cite{blora}, and Inspiration Tree~\cite{inspiration_tree}. For DreamBooth, we train two separate models for content and style using distinct prompts: \texttt{"A photo of a $y_c$"} for content and \texttt{"A subject in $y_s$ style"} for style. While UnzipLoRA~\cite{unziplora} is closely related to our work, it lacks a publicly available code, preventing direct comparison.

As shown in ~\cref{fig:5_qualitative_comparisons}, our method excels at preserving content characteristics while realistically adapting them to new environments. In contrast, existing methods struggle with overfitting or content misalignment. For example, DreamBooth strongly overfits to style, often failing to follow the target prompt. B-LoRA and Inspiration Tree, while capable of transferring content, produce suboptimal results when recontextualizing content into different environments. In terms of style alignment, other methods exhibit content leakage, where style representations unintentionally retain content details. Notably, in the fox and bear examples (left, first and second row), other methods fail to fully separate content and style, causing unwanted subject appearances in stylized outputs. Our approach mitigates this issue, achieving faithful stylization without content artifacts.

\begin{table}[t]
    \small
    \centering
    \setlength{\tabcolsep}{2.5pt} 
    \begin{tabular}{cccccccc}
        \toprule
         &  &  & \multicolumn{2}{c}{\bf Content Align} & \multicolumn{2}{c}{\bf Style Align} & \bf Text Align \\
        \cmidrule(lr){4-5} \cmidrule(lr){6-7} \cmidrule(lr){8-8} 
         \bf SA & \bf SVD & \bf KV & \bf CSD-C & \bf CLIP-I & \bf CSD-S & \bf DINO & \bf CLIP-T \\
        \midrule
        
        \checkmark & \checkmark & \checkmark & \textbf{0.603} &\textbf{0.751} & \textbf{0.564} &\textbf{0.517} & \textbf{0.330}\\
        \midrule
        \checkmark  &\checkmark  & & 0.581 & 0.702 & 0.559 & 0.509 & 0.315 \\
        \checkmark &&\checkmark & 0.601 & 0.725 & 0.503 & 0.422 & 0.289\\
        &\checkmark & \checkmark & 0.501 & 0.612 & 0.547 & 0.508 & 0.270 \\
        &  &  &0.482 & 0.527&0.431&0.320&0.302 \\
        \bottomrule
    \end{tabular}
    \vspace{-5pt}
    \caption{\bf Effectiveness of each component.}
    \label{tab:5_ablate_components}
    \vspace{-15pt}
\end{table}

\myheading{Quantitative comparison.} \cref{tab:5_compare_sota} presents a quantitative comparison of our method with existing approaches in the CSD-100 dataset. Our approach achieves the highest scores in both content alignment and style alignment across backbones, demonstrating superior content identity preservation and faithful stylization. While DreamBooth-C attains high content alignment and DreamBooth-S achieves strong style alignment, both exhibit lower text alignment scores, indicating overfitting to the input image. In contrast, our method maintains the highest text alignment score, reflecting better generalization in following textual descriptions.

\myheading{User Study.}
In our user study, participants were shown an input image alongside the outputs of two competing methods. Each output group included two images for content recontextualization and two for style transfer. Participants evaluated and selected the superior output group based on five criteria: Image Quality, Prompt Adherence, Content Alignment, Style Alignment, and Overall Quality.   
The study gathered 7,500 responses from 100 participants. 

As shown in \cref{fig:userstudy}, our method significantly outperforms others in content and style alignment. Additionally, it achieves the highest preference in prompt adherence, demonstrating its effectiveness in preserving both textual guidance and visual fidelity after content-style decomposition. Further details are provided in the Supp.

\subsection{Ablation Study}

\begin{figure}[t!]
	\centering
	\includegraphics[trim={0 0 0 0},clip,width=\linewidth]{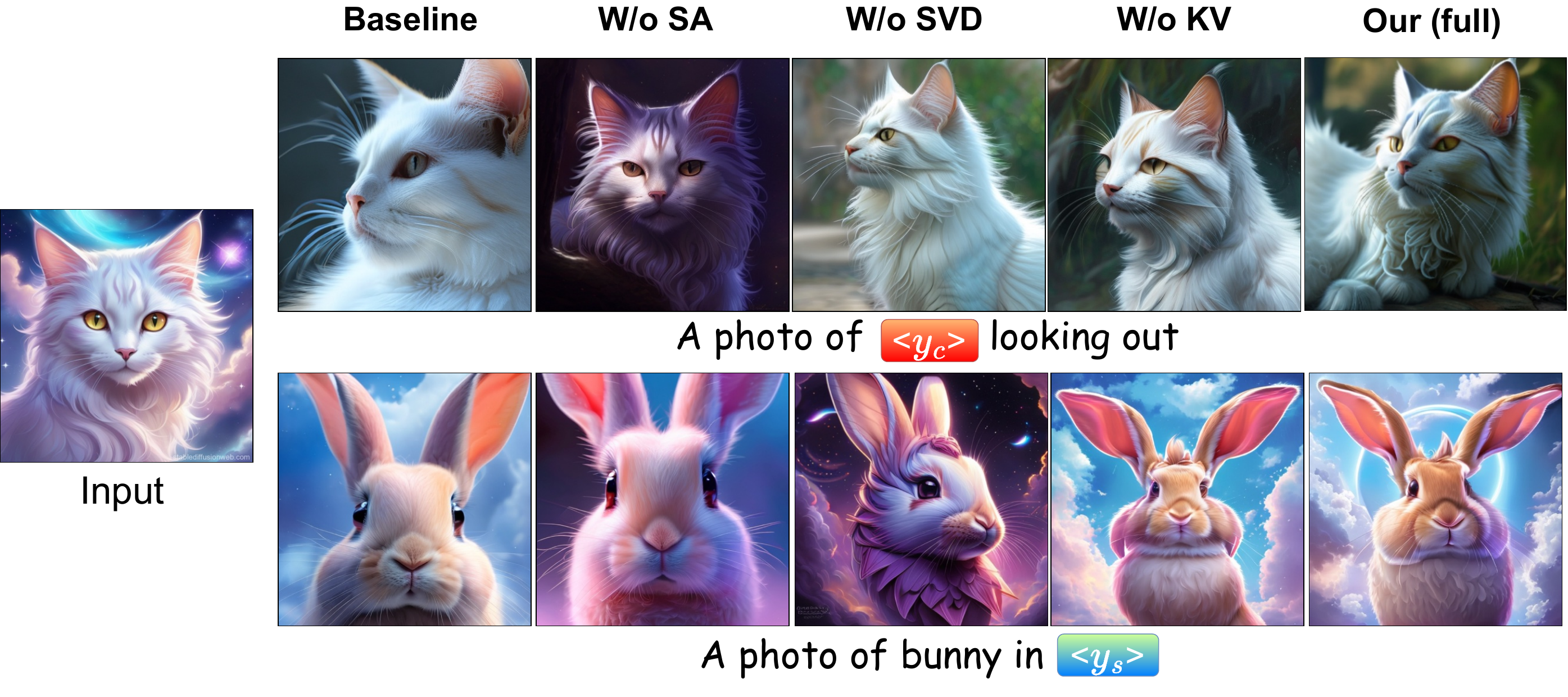}
        \vspace{-20pt}
	\caption{\bf Illustrative effectiveness of each component.}
	\label{fig:5_qualitative_ablation}
    \vspace{-10pt}
\end{figure}

\myheading{Effectiveness of each component} is presented in ~\cref{tab:5_ablate_components} and \cref{fig:5_qualitative_ablation}. Removing the scale-aware strategy severely impacts the model’s ability to separate content from style, leading to the lowest content-alignment and text-alignment scores. Without rectification, style alignment degrades due to content leakage in the generated images. Additionally, eliminating augmented K-V memories weakens the model’s ability to capture both content and style, resulting in lower scores across multiple metrics.

\begin{table}[t]
\small
    \centering
    \setlength{\tabcolsep}{2.5pt} 
    \begin{tabular}{lcccccc}
        \toprule
         & & \multicolumn{2}{c}{\bf Content Align} & \multicolumn{2}{c}{\bf Style Align} & \bf Text Align \\
        \cmidrule(lr){3-4} \cmidrule(lr){5-6} \cmidrule(lr){7-7} \bf Blocks
        & \bf \#Params & \bf CSD-C & \bf CLIP-I & \bf CSD-S & \bf DINO& \bf CLIP-T \\
        \midrule
        First & \textbf{7K} & 0.603 & \textbf{0.751} &0.564 &0.517 & \textbf{0.330}  \\
        Half & 115K  & 0.611  & 0.732 & \textbf{0.582}  & \textbf{0.550}  & 0.300  \\
        All & 230K  & \textbf{0.621}  & 0.733 & 0.580  & 0.541  & 0.292  \\
        \bottomrule
    \end{tabular}
    \vspace{-5pt}
    \caption{\bf Study on \# blocks to apply augmented K-V memory.}
    \vspace{-10pt}
    \label{tab:5_ablate_blocks}
\end{table}

\myheading{Study on the number of blocks to apply augmented K-V memories.} We evaluate three settings: applying K-V only to the first block, to the first half of all blocks (15 blocks), and across all 30 transformer blocks. As shown in~\cref{tab:5_ablate_blocks}, increasing the number of blocks improves content and style alignment but slightly reduces text alignment, likely due to overfitting, while providing only marginal gains relative to the added computational cost. Given this trade-off, we choose to apply K-V to a single block, as it offers the best balance between efficiency and performance.

\myheading{Study on SVD-based style rectification.}  
We empirically evaluate different $r$ in~\cref{tab:5_choice_rectify} when selecting the top $r$ components from a total of 200, and found $r = 10$ is optimal.  

\begin{table}[t]
\small
    \centering
    \setlength{\tabcolsep}{5pt} 
    \begin{tabular}{lccccc}
        \toprule
        \bf  & \multicolumn{2}{c}{\bf Content Align} & \multicolumn{2}{c}{\bf Style Align} & \bf Text Align \\
        \cmidrule(lr){2-3} \cmidrule(lr){4-5} \cmidrule(lr){6-6}
        \bf Setting & \bf CSD-C & \bf CLIP-I & \bf CSD-S & \bf DINO & \bf CLIP-T \\
        \midrule
        Top 3 &0.581 &0.726 &0.561 &0.524 &0.289 \\
        Top 5 &0.598 &0.724 &0.563 &\textbf{0.528} &0.317 \\
        Top 10 & 0.603 &\textbf{0.751} & \textbf{0.564} &0.517 &\textbf{0.330} \\
        Top 20 &\textbf{0.605} &0.723 &0.561 &0.512 &0.302 \\
        \bottomrule
    \end{tabular}
    \vspace{-7pt}
    \caption{\bf Study on \# ranks in SVD-based rectification.}
    \label{tab:5_choice_rectify}
    \vspace{-10pt}
\end{table}

\myheading{Study on \# tokens for learning content and style.}  
As shown in \cref{tab:5_ablate_number_of_embedding}, increasing \# tokens does not consistently improve identity preservation. While a moderate increase (4 tokens) enhances content alignment, using too many tokens (16) introduces artifacts and reduces alignment. Based on these findings, we select 4 tokens as the optimal choice.

\begin{table}[t]
    \small
    \centering
    \setlength{\tabcolsep}{5pt} 
    \begin{tabular}{cccccc}
        \toprule
        \bf  & \multicolumn{2}{c}{\bf Content Align} & \multicolumn{2}{c}{\bf Style Align} & \bf Text Align \\
        \cmidrule(lr){2-3} \cmidrule(lr){4-5} \cmidrule(lr){6-6}
        \bf \# Tokens & \bf CSD-C & \bf CLIP-I & \bf CSD-S & \bf DINO & \bf CLIP-T \\
        \midrule
        1 &0.581 &0.701 &0.544 &0.470 &0.316 \\
        2 &0.593 &0.713 &0.552 &0.473 & 0.316 \\
        4 &\textbf{0.603} &\textbf{0.751} &0.564 &0.517 & \textbf{0.330} \\
        8 &0.601 &0.715 &0.550 &0.516 &0.314 \\
        16 &0.572 &0.683 &\textbf{0.582}& \textbf{0.534} &0.301 \\
        \bottomrule
    \end{tabular}
    \vspace{-7pt}
    \caption{\bf Study on \# tokens used for learning content and style.}
    \label{tab:5_ablate_number_of_embedding}
    \vspace{-10pt}
\end{table}

\section{Conclusion}
We have introduced CSD-VAR, a novel method for decomposing a single image into separate content and style representations using a Visual Autoregressive model, along with CSD-100 as a benchmark dataset for this task. By analyzing scale-level details, we have proposed a scale-aware alternating optimization strategy to enhance content-style disentanglement. Additionally, we have introduced SVD-based style embedding to mitigate content leakage and augmented K-V memories to improve subject identity preservation. Our experiments show that CSD-VAR outperforms existing methods, establishing it as an effective approach for controllable text-to-image generation and creative exploration from a single image.

\myheading{Discussion.}  
While our method achieves significant improvements, it struggles with images containing subjects with intricate details, highlighting the need for better disentanglement and fine-grained representation learning.  
In future work, we aim to refine our approach to better capture such details and expand CSD-100 beyond an evaluation benchmark, exploring its potential as a training dataset for learning-based content-style decomposition methods.

{
    \small
    \bibliographystyle{ieeenat_fullname}
    \bibliography{main}
}

\newpage

\clearpage
\appendix

\section{Appendix}

This appendix includes our supplementary materials as follow:
\begin{itemize}
    \item Validation Data in~\cref{appendix:eval_data}
    \item CSD-100 Dataset Analysis in~\cref{appendix:data_analysis}
    \item Additional Qualitative Results in~\cref{appendix:add_qualitative}
    \item More ablation experimental settings in~\cref{appendix:more_ablate}
    \item User Study in~\cref{fig:appendix_user_study}
\end{itemize}

\subsection{Validation Data}  
\label{appendix:eval_data}  

\begin{figure}[t!]  
	\centering  
	\includegraphics[width=\linewidth]{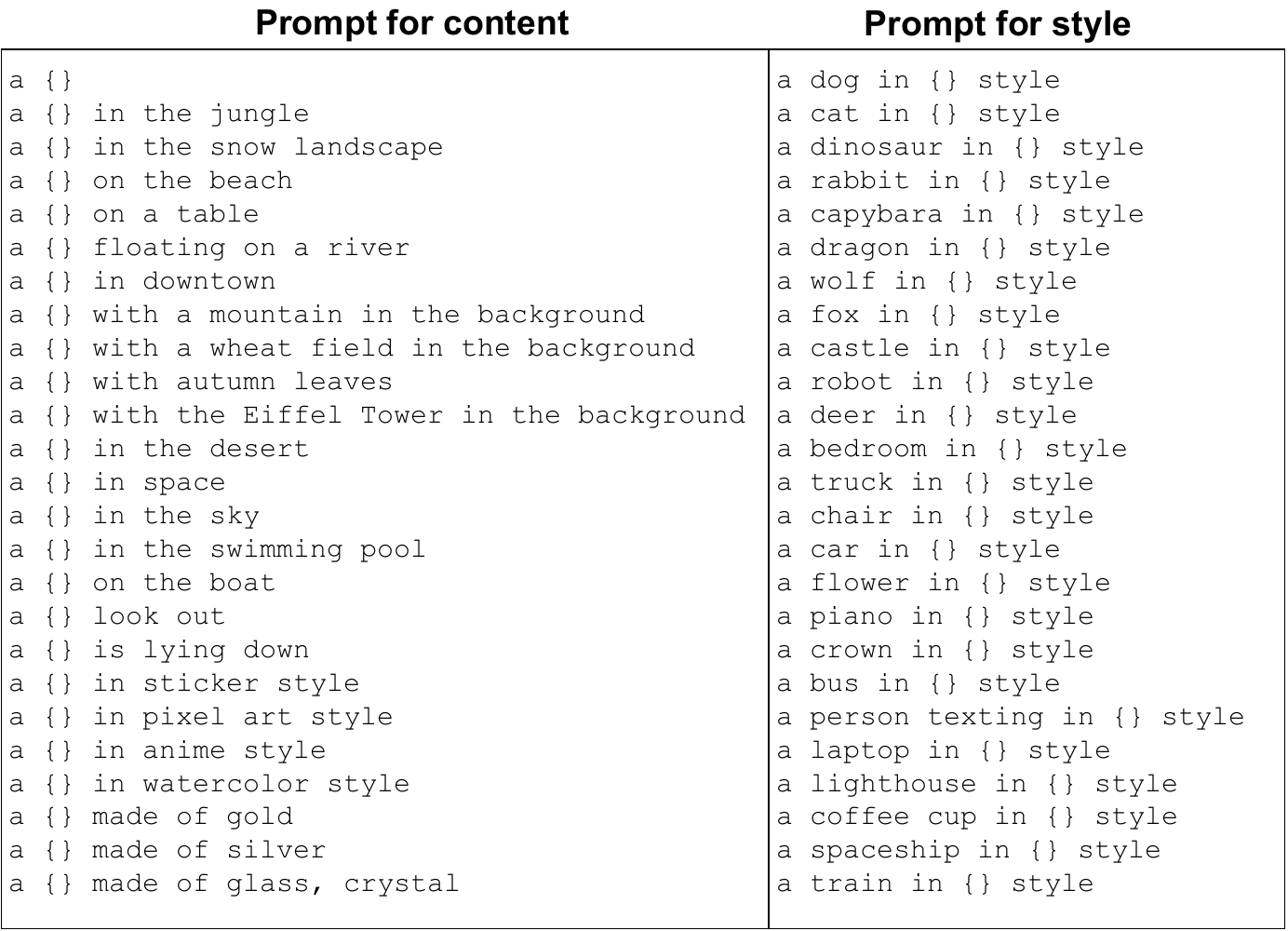}  
	\caption{List of 50 prompts used in our experiments for evaluating content-style decomposition.}  
	\label{fig:appendix_prompt}  
        \vspace{-10pt}
\end{figure}

\begin{figure*}[t!]  
	\centering  
	\includegraphics[width=\linewidth]{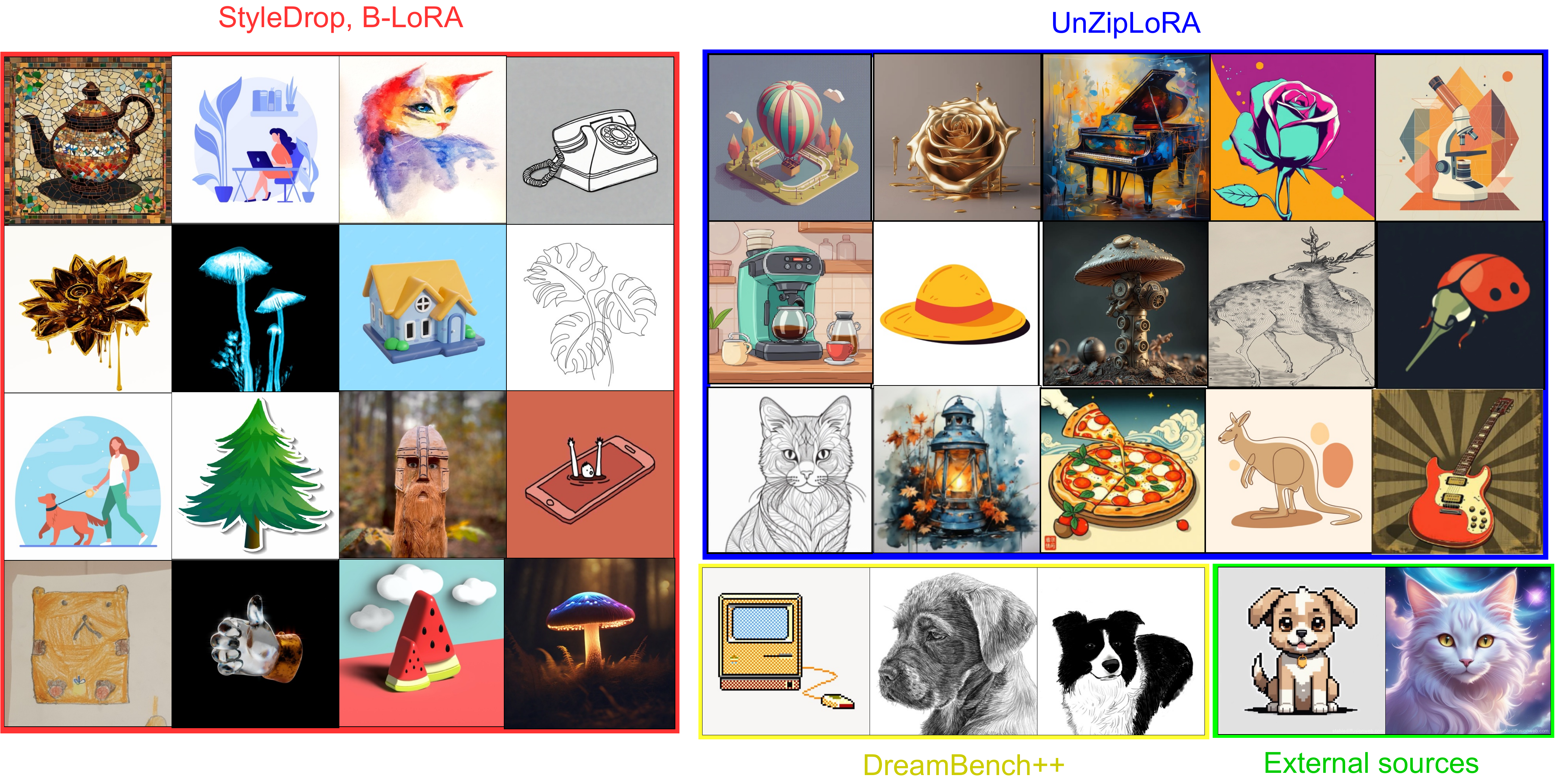}  
	\caption{Overview of the validation dataset, consisting of 35 curated concepts sourced from existing personalization benchmarks.}  
	\label{fig:appendix_valid_data}  
\end{figure*}

To assess the effectiveness of our scale-wise detail analysis and ablation studies, we construct a validation dataset comprising 35 diverse concepts. These concepts are primarily curated from StyleDrop~\cite{styledrop}, B-LoRA~\cite{blora}, UnZipLoRA~\cite{unziplora}, and DreamBench++~\cite{dreambenchplus}. This dataset allows us to systematically validate our assumptions regarding hierarchical scale representations and evaluate different experimental configurations. \cref{fig:appendix_valid_data} provides an overview of the validation dataset, while \cref{fig:appendix_prompt} presents the set of 50 prompts used across all experiments.

\begin{figure}[t!]  
	\centering  
	\includegraphics[width=\linewidth]{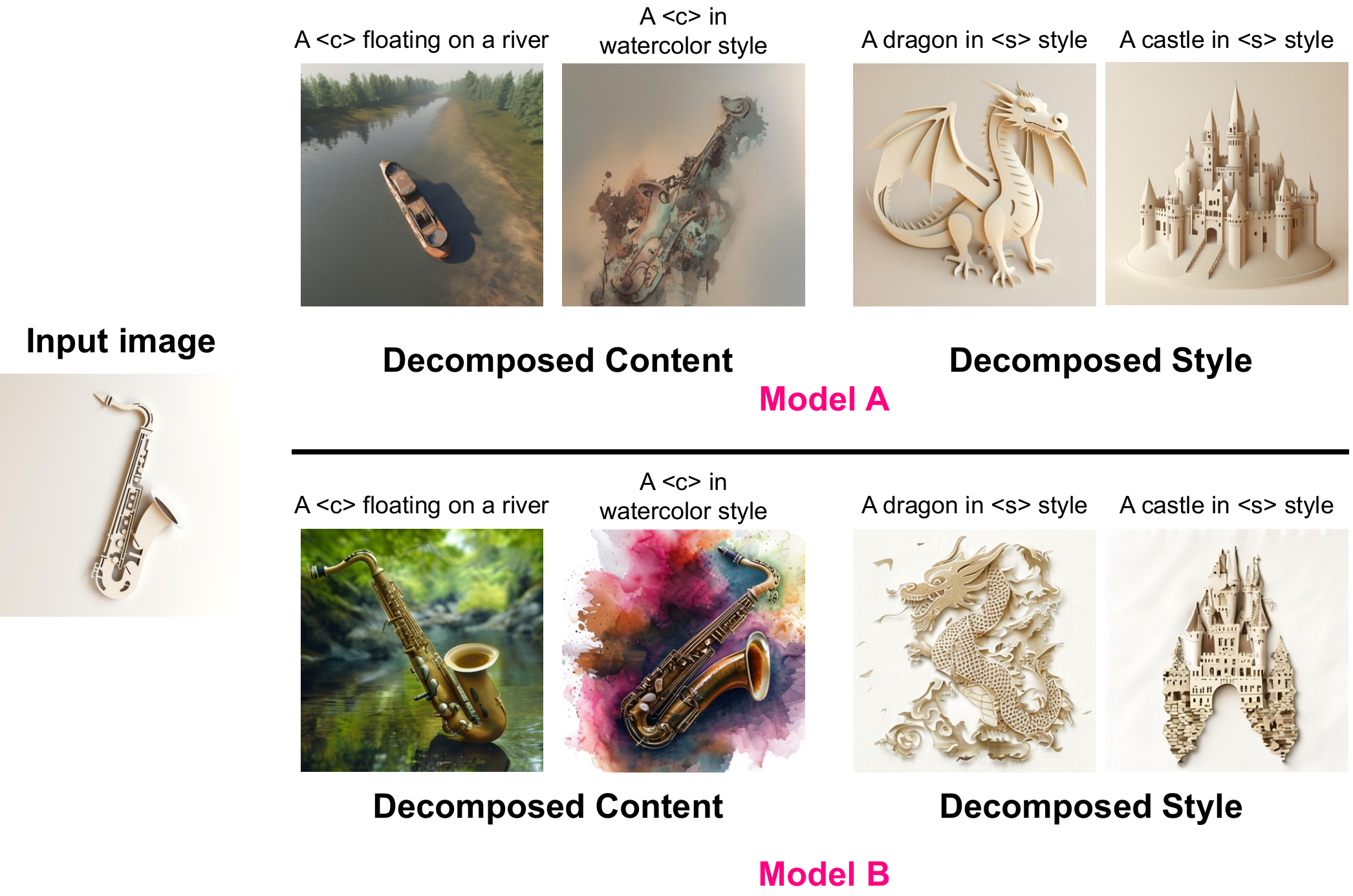}  
	\caption{Our user study interface}  
	\label{fig:appendix_user_study}  
\end{figure}

\subsection{CSD-100}
\label{appendix:data_analysis}  
\begin{figure*}[t!]
	\centering	\includegraphics[width=\linewidth]{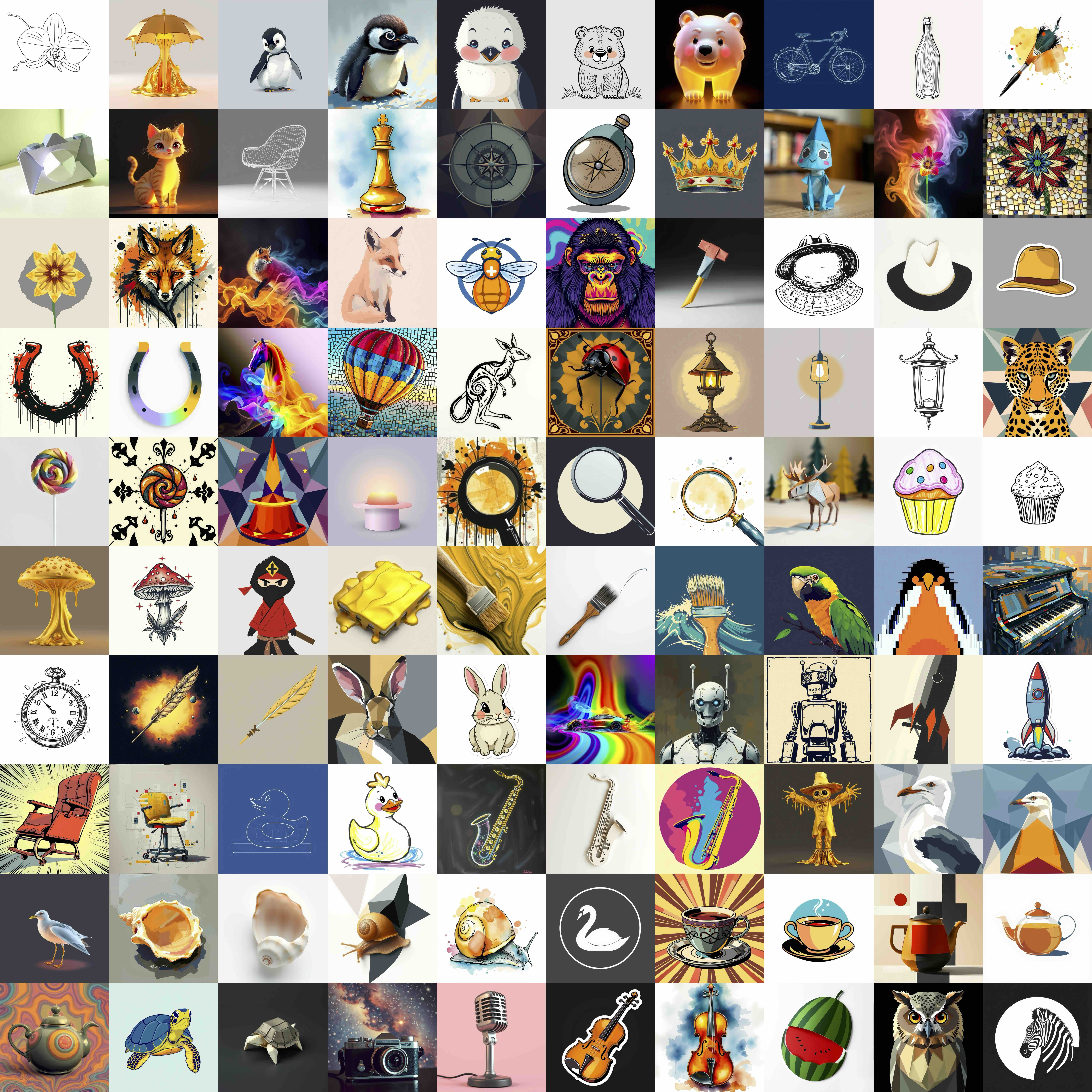}
	\caption{Visualization of the full CSD-100 dataset, showcasing its diverse content and style pairings}
	\label{fig:full_dataset_csd}
\end{figure*}

\begin{figure*}[t!]
	\centering	\includegraphics[width=\linewidth]{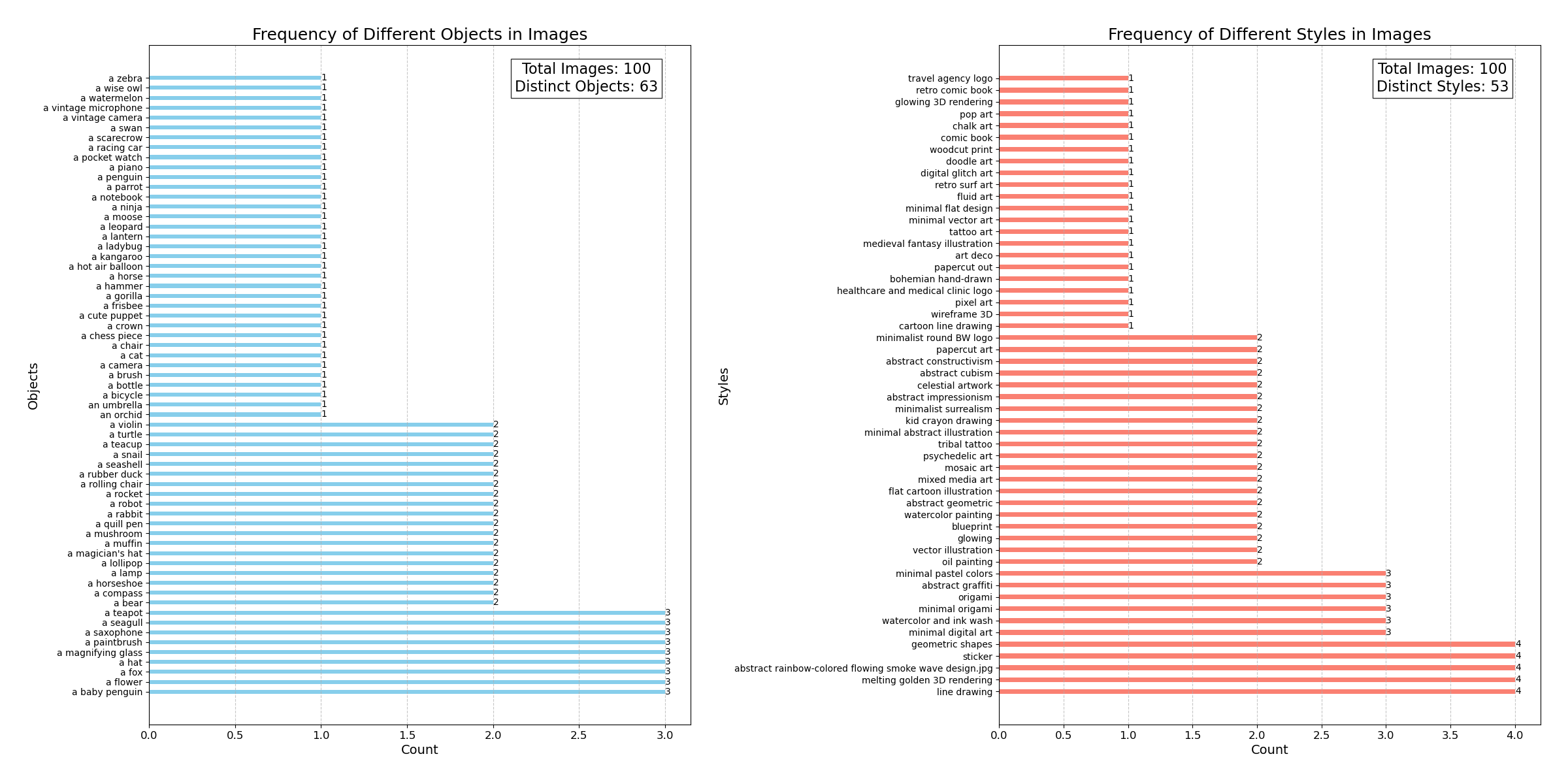}
	\caption{Distribution of content and style in the CSD-100 dataset}
	\label{fig:analysis-dataset}
\end{figure*}

The CSD-100 dataset in~\cref{fig:full_dataset_csd} is designed to provide a comprehensive benchmark for content-style decomposition (CSD) models.\cref{fig:analysis-dataset} illustrates the content and style distributions in the CSD-100 dataset. The dataset comprises 63 distinct objects (content) and 53 unique styles, providing a diverse set of content-style pairs for evaluating decomposition models.
\cref{fig:analysis-dataset} (Left) depicts the frequency distribution of content categories, where most content types appear only once or twice, ensuring a broad variety of objects. \cref{fig:analysis-dataset} (Right) shows the distribution of styles, with certain styles occurring more frequently, particularly those with well-defined characteristics (e.g., line drawing, geometric shapes, abstract digital art).
By balancing content and style diversity while maintaining a representative distribution, CSD-100 serves as a well-curated benchmark for assessing the effectiveness of content-style decomposition models.

\subsection{Additional Qualitative Results}
\label{appendix:add_qualitative}
\begin{figure*}[t!]  
	\centering  
	\includegraphics[width=\linewidth]{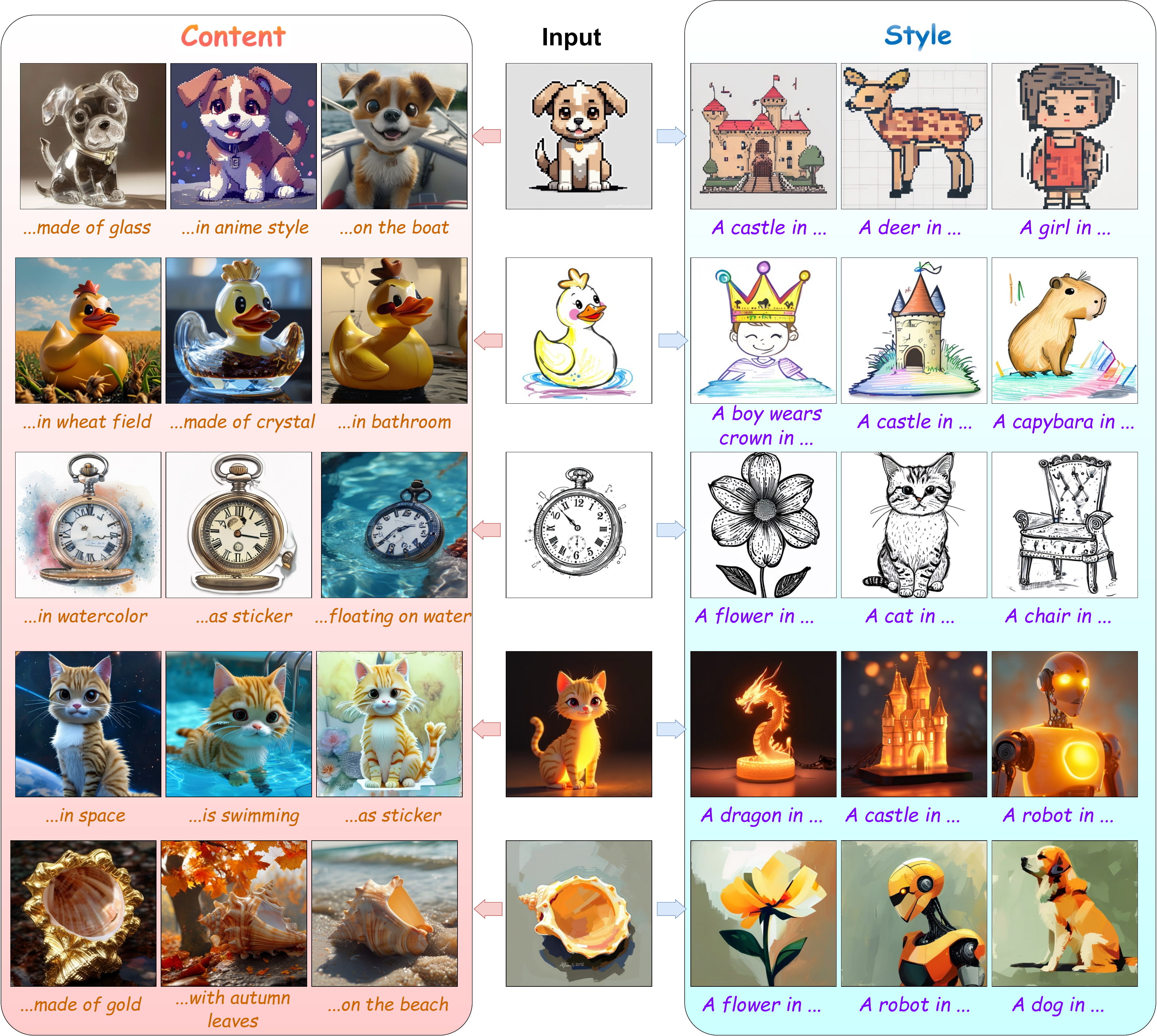}  
	\caption{Additional qualitative results from our CSD-VAR}  
	\label{fig:additional_results}  
\end{figure*}

We present additional qualitative results of our method, as shown in~\cref{fig:additional_results}

\subsection{More ablation experimental settings}
\label{appendix:more_ablate} 

\myheading{Study on the Number of Augmented K-V Memories.}  
The results in \cref{tab:5_ablation_kv} show that increasing the number of K-V memories does not consistently improve performance across all metrics. We hypothesize that adding more K-V pairs complicates model distribution alignment, leading to diminishing returns. As a result, we adopt a single K-V memory as the optimal setting.

\begin{table}[t]
\footnotesize
    \centering
    \setlength{\tabcolsep}{3pt} 
    \begin{tabular}{cccccc}
        \toprule
        \bf  & \multicolumn{2}{c}{\bf Content Align} & \multicolumn{2}{c}{\bf Style Align} & \bf Text Align \\
        \cmidrule(lr){2-3} \cmidrule(lr){4-5} \cmidrule(lr){6-6}
        \bf \# Augmented $K, V$ & \bf CSD-C & \bf CLIP-I & \bf CSD-S & \bf DINO & \bf CLIP-T \\
        \midrule
        1 &0.603 &\textbf{0.751} &\textbf{0.564} &0.517 &\textbf{0.330} \\
        2  &\textbf{0.608} &0.724 &0.560 &\textbf{0.521} &0.320 \\
        5 &0.607 &0.732 &0.560 &0.517 &0.318 \\
        \bottomrule
    \end{tabular}
    \caption{Ablation study on the number of augmented K-V memories used in self-attention.}
    \label{tab:5_ablation_kv}
    \vspace{-10pt}
\end{table}

\subsection{User Study}
We conducted a user study with 100 participants to compare our model against an alternative method. Each participant answered 15 questions, selecting their preferred output based on five multiple-choice criteria. For each question, users evaluated two examples per output group, focusing on content and style (see~\cref{fig:appendix_user_study}). To ensure fairness, model outputs were anonymized and their order randomized.

\end{document}